\documentclass[letterpaper,journal]{IEEEtran}

\usepackage{amsmath,amssymb,amsfonts}
\usepackage{graphicx}
\usepackage{booktabs}
\usepackage{array}
\usepackage{microtype}
\usepackage{url}
\usepackage{cite}
\usepackage{algorithm}
\usepackage[noend]{algpseudocode}
\usepackage{placeins}
\usepackage{needspace}
\usepackage{textcomp}
\usepackage{stfloats}
\usepackage[colorlinks=true,linkcolor=blue,citecolor=blue,urlcolor=blue]{hyperref}

\newcommand{\method}{LARC}
\newcommand{\FO}{\mathrm{FO}}
\newcommand{\Rhat}{\widehat{\mathcal{R}}}
\newcommand{\Ob}{\mathcal{O}}

\title{LARC: Lazy Adaptive Reachability Certification of Robot Manipulator Trajectories}

\author{Yu Feng,~\IEEEmembership{Member,~IEEE}, Hao Wu,~\IEEEmembership{Student Member,~IEEE}, Yuzhe Wang, and
Jianshu Zhou,~\IEEEmembership{Member,~IEEE}%
\thanks{This work was supported in part by the National University of
Singapore under the NUS Start-up Grant (FY2026).}%

\thanks{Yu Feng, Hao Wu, and Jianshu Zhou are with the Department of Mechanical
Engineering, National University of Singapore, Singapore. Corresponding author:
Jianshu Zhou (e-mail: jianshuzhou@nus.edu.sg).}
\thanks{Yuzhe Wang is with Singapore Institute of Manufacturing Technology, Agency for Science Technology and Research, Singapore.}
}

\begin{document}
\maketitle

\begin{abstract}
Checking a trajectory at sampled states can miss collisions between them.
Reachability-based certification bounds the motion between states, but a uniform time
partition wastes computation where clearance is ample. In this study, we report lazy adaptive
reachability certification (\method), which checks a planned trajectory and bisects only
the intervals whose clearance test is inconclusive. For piecewise-cubic Hermite joint
trajectories, \method{} places each link's capsule at the interval midpoint and inflates it
by an upper bound on the link displacement over half the interval, computed from exact
per-joint speed maxima. On 160 simulated AgileX PIPER
trajectories from 80 start--goal pairs, \method{} matched every decision of the fixed-fine
baseline at depth nine with 20328 interval evaluations (24.8\% of the baseline) and a median
paired speedup of $10.28\times$. A MoveIt/FCL audit of 158051 sampled states found
collisions in 21 direct-interpolation controls. \method{} certified none of them and
rejected 27 of the 139 trajectories that passed the audit.
The feasibility of LARC has been illustrated by A physical PIPER demonstration reaching among cardboard obstacles. This study provide a new method for robotic trajectory safety planning.
\end{abstract}


\section{Introduction}
\IEEEPARstart{R}{obotic} manipulation needs reliable contact at the hand and safe motion of
the arm. On the contact side, tactile sensors measure contact forces and surface
interactions~\cite{kappassovTactileSensingDexterous2015,andrussowMinsightFingertipSized2023,fengSoftMechanoluminescentSkin2026,fengBioinspiredSoftMagnetociliary2026,wuSpectraTacCompactCameraFree2026},
force feedback is used to adjust grasps~\cite{fengMagCilia2026}, and multimodal sensing
supports learned in-hand manipulation~\cite{huangDIHTele2025}. Compliant grippers and
dexterous hands adapt mechanically to object
geometry~\cite{mengaldoConciseGuideModelling2022,deimelNovelTypeCompliant2016a,zhouEverythingGraspingGripperUniversal2025,zhouDexterousCompliantDexCo2025b,wuDexLinkHandCompact2026,wuSyLinkHandSynergyInspired2026,caoOrigamiGripper2024},
and variable-stiffness wrists adjust orientation and compliance during
manipulation~\cite{zhouBioinspiredSoftWrist2023}. These capabilities become relevant once the
hand reaches the object, and bringing it there is the task of the arm. Arm reach and
positioning have also been studied directly: confined workspaces have motivated manipulators
that adapt their reach and configuration~\cite{zhouPrismaticBending2025}, and manipulation
interfaces can pair large-range arm positioning with precise local
control~\cite{zhouGlobalLocal2025}.

Safe arm motion concerns every link, not only the gripper. When an arm reaches into a shelf,
for example, the gripper may pass cleanly through the opening while the elbow approaches the
shelf edge. Planners usually return such a motion as a sequence of waypoints, and the standard
check tests every link at each waypoint. This check confirms clearance at the sampled instants
but provides no information about the motion between them, which is precisely when the elbow
nears the shelf. Denser sampling shortens these gaps but cannot close them. A guarantee for
the entire motion therefore requires bounds on the positions that each link can occupy
between samples.

\begin{figure*}[t]
  \centering
  \includegraphics[width=0.86\textwidth,trim=0mm 3.3mm 0mm 1.5mm,clip]{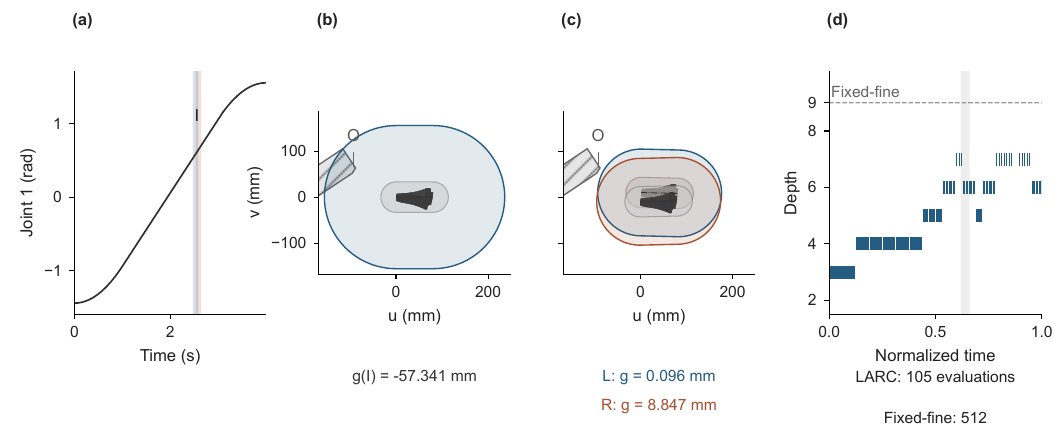}
  \caption{Recorded shelf trajectory. (a) Joint-1 Hermite motion; interval $I$ is bisected
into a blue left half and an orange right half. (b,c) The link--obstacle pair that limits the
parent interval, before and after bisection, in common metric coordinates: collision mesh in
dark gray, midpoint capsule in light gray, colored motion envelopes, and hatched obstacle
$O$. The parent test is inconclusive; both children pass the test over all links. (d) All 53
terminal intervals against $t/T_h$, with $I$ shaded. The dashed line marks the fixed-fine
depth of nine; \method{} needs 105 evaluations instead of 512.}
  \label{fig:overview}
\end{figure*}

Reachability-based certification provides such bounds. It encloses the volume that each link
sweeps over a time interval. Because the enclosure contains every position the link occupies
in that interval, the link keeps the required clearance whenever the enclosure does, and a
single distance test covers the whole interval. Forward-occupancy methods use this property to
design
trajectories~\cite{holmesReachableSetsSafe2020a,michauxSafePlanningArticulated2024a,kwonConformalizedReachableSets2025}.
In this work, analytic enclosures serve a different purpose: the certification of a trajectory
that has already been planned. Fig.~\ref{fig:overview} shows how local enclosure tests on a
recorded joint trajectory produce a time partition.

The tightness of the enclosure depends on the interval length. Over a long interval, the link
sweeps a large region, and the enclosure can intrude into the clearance margin even when the
link itself remains clear; the test is then inconclusive. Shorter intervals yield tighter
enclosures but require more tests, and a uniformly fine partition incurs this cost along the
entire trajectory, including long segments that a single coarse interval would already
certify. Refinement is therefore most useful where the coarse test fails.

\method{} restricts refinement to these intervals. It first tests the entire trajectory as a
single interval. An inconclusive interval is bisected and each half is tested in turn, which
generates a binary tree of intervals. Intervals that pass are not subdivided further, so
motion with ample clearance is certified by a few long intervals and deep refinement is
concentrated near obstacles. If an interval is still inconclusive at the maximum depth, the
trajectory is not certified. Our contributions are as follows:
\begin{itemize}
  \item an adaptive interval certifier whose capsule enclosures use exact per-joint maxima of
  the Hermite joint speeds;
  \item a proof that a certified trajectory keeps every link at the required clearance from
  static obstacles in continuous time; and
  \item a paired comparison with fixed-resolution certifiers on 160 simulated trajectories, checked
  against a sampled MoveIt/FCL audit.
\end{itemize}

\section{Related Work}
\begin{figure}[t]
  \centering
  \includegraphics[width=0.88\columnwidth]{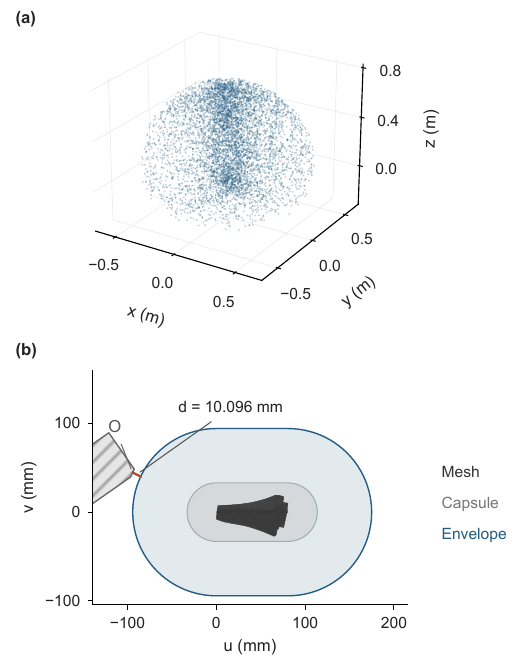}
  \caption{Global workspace and local certification. (a) End-effector-base positions for 6000
joint configurations sampled uniformly within the six official joint limits (seed 20270823).
(b) Terminal interval with the smallest clearance on the shelf
trajectory of Fig.~\ref{fig:overview}. Mesh, capsule, and envelope are projected onto the
plane spanned by the capsule axis and the nearest-obstacle normal, in local metric
coordinates $u,v$; $O$ is the hatched AABB. The envelope--obstacle gap is 10.096~mm;
subtracting $\rho=10$~mm gives $g(I)=0.096$~mm.}
  \label{fig:robotics-context}
\end{figure}

Capability and reachability maps describe the end-effector poses a robot can attain and are
used for base placement, inverse queries, and morphology
analysis~\cite{stulpCombiningAnalysisImitation2009,zachariasUsingModelReachable2009,makhalReuleauxRobotBase2018,kimLearningReachableManifold2021a,rudorferRM4DCombinedReachability2025,cavelliModelingReachabilitySpace2025b,zhuStructuralAnalysisDesign2025a}.
Learned reachability models have been applied to grasp ranking, moving-target tracking, and
embodiment selection for mobile
manipulators~\cite{louLearningGenerate6DoF2020,akinolaDynamicGraspingReachability2021,fengPredictiveReachabilityEmbodiment2025}.
Both classes of model characterize where the end effector can go. A trajectory certificate
instead requires bounds on the space swept by every link during the motion
(Fig.~\ref{fig:robotics-context}).

ARMTD builds reachable sets into online trajectory
design~\cite{holmesReachableSetsSafe2020a}. SPARROWS uses spherical forward occupancy with
exact obstacle distances~\cite{michauxSafePlanningArticulated2024a}, and CROWS combines
learned spherical predictions with conformal buffers~\cite{kwonConformalizedReachableSets2025}.
All three use forward occupancy inside the planner, where the reachable sets shape the
trajectory itself. This work addresses the complementary setting: the trajectory is fixed
before certification, and the problem is how to allocate interval tests along it when the
enclosures are deterministic and analytic.

\section{Problem Formulation}
Consider an $n$-joint serial manipulator following a candidate trajectory $q(t;k)$ over
$t\in[0,T_h]$, where $k$ indexes the candidate. Let $\mathcal{V}_\ell(q)$ denote the
workspace geometry occupied by link $\ell$ at configuration $q$. The forward occupancy of the
link over a time interval $I$ is
\begin{equation}
  \FO_\ell(I;k)=\bigcup_{t\in I}\mathcal{V}_\ell(q(t;k)).
  \label{eq:fo}
\end{equation}
This set contains every point that the link occupies during $I$, including the motion between
recorded waypoints that a waypoint check omits. Because computing the union exactly is
expensive, we construct a computable outer envelope $\Rhat_\ell(I;k)$ such that
\begin{equation}
  \FO_\ell(I;k)\subseteq \Rhat_\ell(I;k).
  \label{eq:containment}
\end{equation}
The external obstacles $\Ob$ are static axis-aligned bounding boxes (AABBs). The required
clearance is a positive margin,
\begin{equation}
 \rho\in\mathbb{R}_{>0}.
 \label{eq:margin}
\end{equation}
In the simulation study, we use $\rho=10$~mm. Writing $d$ for the minimum Euclidean distance between
sets, we define
\begin{equation}
 g(I;k)=\min_{\ell}d\!\left(\Rhat_\ell(I;k),\Ob\right)-\rho.
 \label{eq:certificate}
\end{equation}
Thus, $g$ is the clearance surplus of the most constrained link. An interval passes the clearance
test when $g(I;k)>0$, that is, when every link envelope is more than $\rho$ from the
obstacles; otherwise the test is inconclusive. A trajectory is certified when passing leaf
intervals cover its entire duration.

\section{Lazy Adaptive Reachability Certification}
\subsection{Hermite Trajectories and Speed Bounds}
Each recorded trajectory stores joint positions and velocities at its waypoints, which are the
boundary data of a cubic Hermite segment; we therefore reconstruct the motion as a
piecewise-cubic Hermite curve. On each segment, $q_j(t)$ is cubic and its derivative
$\dot q_j(t)$ is quadratic, so the maximum of $|\dot q_j(t)|$ lies at a clipped-segment
endpoint or at an interior root of $\ddot q_j(t)$. Evaluating these candidates on every
segment that intersects $I$ gives the exact bound
\begin{equation}
 \bar\omega_j(I)=\max_{t\in I}|\dot q_j(t)|.
 \label{eq:speed-bound}
\end{equation}
Computing the maximum exactly keeps this bound tight, at the cost of a few polynomial
evaluations per segment.

\subsection{Midpoint Capsules and Motion Inflation}
We model the arm with eleven capsules fitted around the official PIPER collision meshes with
1~mm of padding. Capsules are well suited to this purpose because their distance to a box
reduces to a segment--box distance, which is inexpensive to compute. The envelope is obtained
by evaluating the capsule at the interval midpoint and dilating it by the largest distance
that any of its points can travel within half the interval. For $I=[t_m-h,t_m+h]$, let
$C_\ell(q(t_m))$ denote the capsule of link $\ell$ at the interval midpoint. The motion of a
point on link $\ell$ is induced by the joints between the base and the link, and joint $j$
contributes a speed of at most $a_{\ell j}\bar\omega_j(I)$, where the calibrated constant
$a_{\ell j}$ bounds the distance from the joint axis to any capsule point. Summing over these
joints and multiplying by the half-width $h$ gives the interval motion radius
\begin{equation}
 b_\ell(I)=h\sum_{j=1}^{n}a_{\ell j}\bar\omega_j(I).
 \label{eq:motion-radius}
\end{equation}
Dilating the midpoint capsule by this radius yields the occupancy envelope
\begin{equation}
 \Rhat_\ell(I;k)=C_\ell(q(t_m))\oplus\mathbb{B}(0,b_\ell(I)).
 \label{eq:envelope}
\end{equation}
Here $\oplus$ denotes the Minkowski sum and $\mathbb{B}(0,b)$ a ball of radius $b$. The
capsule accounts for the shape of the link and the added radius for its motion over the
interval. In the implementation, the segment--AABB distance is compared with the capsule
radius plus $b_\ell(I)+\rho$, which is equivalent to the test in~\eqref{eq:certificate}.

\subsection{Adaptive Certificate Tree}
The search is a lazy bisection (Algorithm~\ref{alg:larc}). Starting from the full interval
$[0,T_h]$, the tree is traversed depth first. A node with positive clearance becomes a
certified leaf and is not revisited; any other node is bisected, and the recursion continues
until the test passes or the maximum depth $d_{\max}=9$ is reached. A single inconclusive leaf
at that depth leaves the trajectory uncertified. The search nonetheless evaluates both
branches of every split, so that the stored leaves cover the whole motion and identify where
certification failed.

\begin{algorithm}[t]
\caption{Lazy adaptive certification of one trajectory}
\label{alg:larc}
\small
\begin{algorithmic}[1]
\Require Trajectory $k$, horizon $[0,T_h]$, obstacles $\Ob$, margin $\rho$, maximum depth $d_{\max}$
\Function{Certify}{$I,d$}
  \State construct $\Rhat_\ell(I;k)$ and evaluate $g(I;k)$
  \If{$g(I;k)>0$}
    \State record certified leaf $I$; \Return \textsc{True}
  \ElsIf{$d=d_{\max}$}
    \State record inconclusive leaf $I$; \Return \textsc{False}
  \Else
    \State split $I$ into $I_L,I_R$
    \State $c_L\leftarrow\Call{Certify}{I_L,d+1}$
    \State $c_R\leftarrow\Call{Certify}{I_R,d+1}$
    \State \Return $c_L\land c_R$
  \EndIf
\EndFunction
\State \Return \Call{Certify}{$[0,T_h],0$}
\end{algorithmic}
\end{algorithm}

\subsection{Fixed-Resolution Baselines}
To quantify the benefit of adaptivity, we compare \method{} with two fixed partitions that
share all other components: the trajectories, capsules, speed bounds, obstacles, distance
evaluation, and the 10~mm clearance margin. Fixed-coarse evaluates all 16 uniform intervals
at depth four, and fixed-fine all 512 intervals at depth nine. \method{} refines to at most the
depth of fixed-fine but stops splitting an interval once it passes, so the comparison shows
how much of the fine partition is required. Fixed-fine also serves as the reference for
certificate decisions, and a sampled MoveIt/FCL audit as the reference for collisions.

\section{Safety Analysis}
The guarantee follows almost directly from the construction. The argument rests on
containment: if an outer envelope keeps the required clearance from an obstacle, so does the
link geometry inside it. We assume that the capsules contain the link
geometry, that the motion bounds ensure~\eqref{eq:containment}, and that the external
obstacles stay fixed during the certified horizon.

\noindent\textbf{Theorem 1 (continuous-time external-obstacle clearance).}
Suppose the leaf intervals $\{I_r\}$ cover $[0,T_h]$ and satisfy the containment
condition for every link. If $g(I_r;k)>0$ on every leaf, each link remains more than
$\rho$ from the static obstacles $\Ob$ throughout the horizon.

\emph{Proof.} Choose any time $t\in[0,T_h]$ and a leaf interval $I_r$ containing it.
For each link, containment gives
$\mathcal{V}_\ell(q(t;k))\subseteq\Rhat_\ell(I_r;k)$.
The positive certificate implies that $d(\Rhat_\ell(I_r;k),\Ob)>\rho$ for every link.
Replacing a set by a subset cannot decrease its distance to $\Ob$,
so $d(\mathcal{V}_\ell(q(t;k)),\Ob)>\rho$.
Since $t$ and $\ell$ were arbitrary, the bound holds for every link over the whole
horizon.

The argument does not depend on how the leaf intervals are obtained, so the same guarantee
holds for the two fixed-resolution certifiers.

\section{Simulation Study}
\subsection{Simulation Model and Trajectory Pipeline}
We simulated the six-degree-of-freedom AgileX PIPER arm with its official electric gripper,
using ROS~2 Humble, MoveIt~2, and the PIPER URDF/SRDF model files on Ubuntu~22.04/WSL2.
Trajectories were generated once and replayed offline, so that all certifiers evaluate
identical motions. MoveIt's validity service checked the start
and goal states for self-collision and collision with the scene. Paths were planned with the
default \texttt{geometric::RRTConnect} planner of the Open Motion Planning Library (OMPL) and
then time-parameterized. Each OMPL trajectory was reconstructed by piecewise-cubic Hermite
interpolation and rechecked on a 50~Hz grid before it was accepted.

\subsection{Dataset and Pairing}
The dataset covers four scenes (Table~\ref{tab:protocol}): open, easy window, medium window,
and easy shelf. Each scene has 20 independent start--goal pairs, 80 in total. For each pair
we generated two trajectories with the same endpoints and duration. The first is planned with
OMPL and represents typical planner output. The second is a direct smoothstep interpolation
with zero endpoint velocities that serves as a control. It is not planned around obstacles,
so some of these paths collide; these cases test whether \method{} certifies any colliding
trajectory.

Obtaining 80 accepted pairs required 115 generation attempts. The 35 failed attempts comprised
nine invalid starts, 16 invalid goals, six planning failures, and four failed Hermite
rechecks.

\begin{table}[t]
\caption{Simulation protocol.}
\label{tab:protocol}
\centering
\footnotesize
\renewcommand{\arraystretch}{1.03}
\begin{tabular}{@{}p{0.29\columnwidth}p{0.66\columnwidth}@{}}
\toprule
Item & Setting \\
\midrule
Main comparison & \method{} versus fixed-fine (depth 9) \\
Coarse baseline & Fixed-coarse (depth 4) \\
Scenes & Open, easy window, medium window, easy shelf \\
Independent unit & Start--goal pair within a scene; 80 total \\
Paired variants & OMPL-planned and direct control; 160 paths \\
Timing repeats & Three per trajectory and method \\
Sampled audit & MoveIt/FCL at 250~Hz plus original waypoints \\
Outcomes & Certification, false-safe cases, evaluations, runtime, rejections \\
\bottomrule
\end{tabular}
\end{table}

\begin{figure}[t]
  \centering
  \includegraphics[width=0.90\columnwidth]{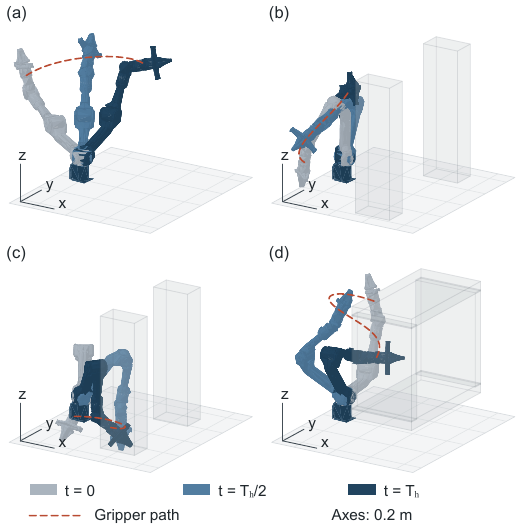}
  \caption{Representative trajectories in the (a) open, (b) easy-window, (c) medium-window,
and (d) easy-shelf scenes; each panel shows the trajectory whose adaptive evaluation count is
closest to the scene median (40 trajectories per scene). Poses at $t=0,T_h/2,T_h$ are drawn
with simplified official PIPER collision meshes. Red dashed curves are
gripper-base paths, and gray boxes are the static AABBs. All panels share one orthographic
view and metric scale; the direction axes are 0.2~m long.}
  \label{fig:scene-replay}
\end{figure}

\subsection{Sampled MoveIt/FCL Audit and Computational Timing}
To evaluate the certificates against an independent reference, we audited every trajectory
with MoveIt and the Flexible Collision Library (FCL), using the official collision geometry
instead of the capsule model. The audit covered a 250~Hz time grid plus all original
waypoints, 158051 states over the 160 trajectories. We call a trajectory
\emph{sampled-clear} if no collision was detected at the audited states. Two types of
disagreement with this reference are possible. An \emph{FCL-detected false-safe} is a
trajectory that \method{} certified although the audit detected a collision, and a
\emph{conservative rejection} is a sampled-clear trajectory that \method{} did not certify.

All three certifiers are implemented in Python. Each ran three times per trajectory in
randomized order (seed 20270824), and we take the median of the three runs to reduce timing
noise. The paired
speedup of a trajectory is its fixed-fine runtime divided by its \method{} runtime. We report
the median of these 160 ratios and the median runtime of each method.

\section{Results}
\subsection{Decision Agreement and Sampled Audit}
Adaptive refinement left every decision unchanged and removed most of the computational cost.
\method{} and fixed-fine certified the same 112 trajectories and agreed on all 160 decisions;
fixed-coarse, with 16 intervals per trajectory, certified 70 (Table~\ref{tab:overall-results}).

The audit detected collisions in 21 direct controls and in none of the 80 OMPL trajectories.
\method{} certified none of these 21 controls, so the dataset contains no FCL-detected
false-safe case. All disagreements with the audit are conservative: 27 of the 139 sampled-clear
trajectories were left uncertified, a conservative-rejection rate of 19.4\%, made up of 22 of
the 80 OMPL trajectories (27.5\%) and 5 of the 59 sampled-clear direct trajectories (8.5\%).

\begin{table}[t]
\caption{Overall results.}
\label{tab:overall-results}
\centering
\scriptsize
\renewcommand{\arraystretch}{1.02}
\begin{tabular}{@{}p{0.34\columnwidth}p{0.19\columnwidth}p{0.37\columnwidth}@{}}
\toprule
Metric & Value & Note \\
\midrule
Independent pairs & 80 & Two trajectories per pair \\
Trajectories & 160 & 40 per scene \\
Certified & 112 & Fine 112; coarse 70 \\
Disagreement with fine & 0/160 & Same certificate model \\
FCL-detected collision & 21 & Direct controls only \\
FCL-detected false-safe & 0/21 & 158051 audited states \\
Conservative rejection & 27/139 (19.4\%) & Of sampled-clear trajectories \\
Interval evaluations & 20328 & Fine 81920; ratio 0.248 \\
Median runtime & 81.8~ms & Fine 801.4~ms \\
Median paired speedup & $10.28\times$ & Median of per-trajectory ratios \\
\bottomrule
\end{tabular}
\end{table}

\subsection{Interval Work and Runtime}
\begin{figure*}[t]
  \centering
  \includegraphics[width=0.79\textwidth,trim=0mm 0.5mm 0mm 2.4mm,clip]{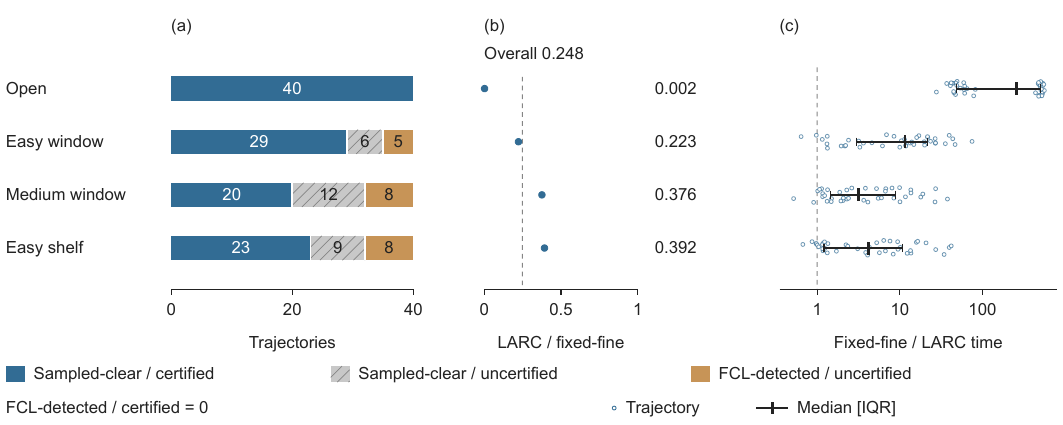}
  \caption{Results for 160 trajectories from 80 start--goal pairs (40 trajectories per scene).
(a) Outcomes against the 250~Hz sampled MoveIt/FCL audit. (b) Summed interval evaluations of
\method{} relative to fixed-fine, by scene; the dashed line is the overall ratio, 0.248. (c)
Fixed-fine to \method{} runtime ratio of each trajectory, on a logarithmic axis. Circles are
individual trajectories; black marks show scene medians and interquartile ranges (IQRs) across
trajectories. The dashed line marks equal runtime.}
  \label{fig:m4-results}
\end{figure*}

The reduction is most directly visible in the number of interval tests. Fixed-fine evaluated 512
intervals per trajectory, 81920 in total, and \method{} needed 20328, a ratio of 0.248, or
75.2\% fewer evaluations (Fig.~\ref{fig:m4-results}). Median runtimes were 81.8~ms for
\method{} and 801.4~ms for fixed-fine, and the median paired speedup was $10.28\times$.

\subsection{Scene Dependence}
The savings depend strongly on the scene (Table~\ref{tab:scene-results};
Fig.~\ref{fig:scene-replay}). In the open scene, where no obstacle lies near the arm, all 40
trajectories passed at the root interval, so \method{} required a single evaluation per
trajectory. This yields an evaluation ratio of 0.002 and a median paired speedup of
$260.31\times$. The
three scenes with obstacles needed further refinement: their evaluation ratios range from
0.223 to 0.392, and their median paired speedups from $3.14\times$ in the medium window to
$11.58\times$ in the easy window.

\begin{table}[t]
\caption{Results by scene. FCL: audit-detected collisions; Cert.: certified;
Rej.: conservative rejections; Eval.: evaluation ratio to fixed-fine;
Speed: median paired speedup. Counts are numbers of trajectories.}
\label{tab:scene-results}
\centering
\scriptsize
\setlength{\tabcolsep}{2pt}
\begin{tabular}{@{}lrrrrrr@{}}
\toprule
Scene & $N$ & FCL & Cert. & Rej. & Eval. & Speed \\
\midrule
Open & 40 & 0 & 40 & 0 & 0.002 & $260.31\times$ \\
Easy window & 40 & 5 & 29 & 6 & 0.223 & $11.58\times$ \\
Medium window & 40 & 8 & 20 & 12 & 0.376 & $3.14\times$ \\
Easy shelf & 40 & 8 & 23 & 9 & 0.392 & $4.18\times$ \\
Overall & 160 & 21 & 112 & 27 & 0.248 & $10.28\times$ \\
\bottomrule
\end{tabular}
\end{table}

\Needspace{7\baselineskip}
\subsection{Hardware Demonstration}
\begin{figure}[!t]
  \centering
  \includegraphics[width=\columnwidth]{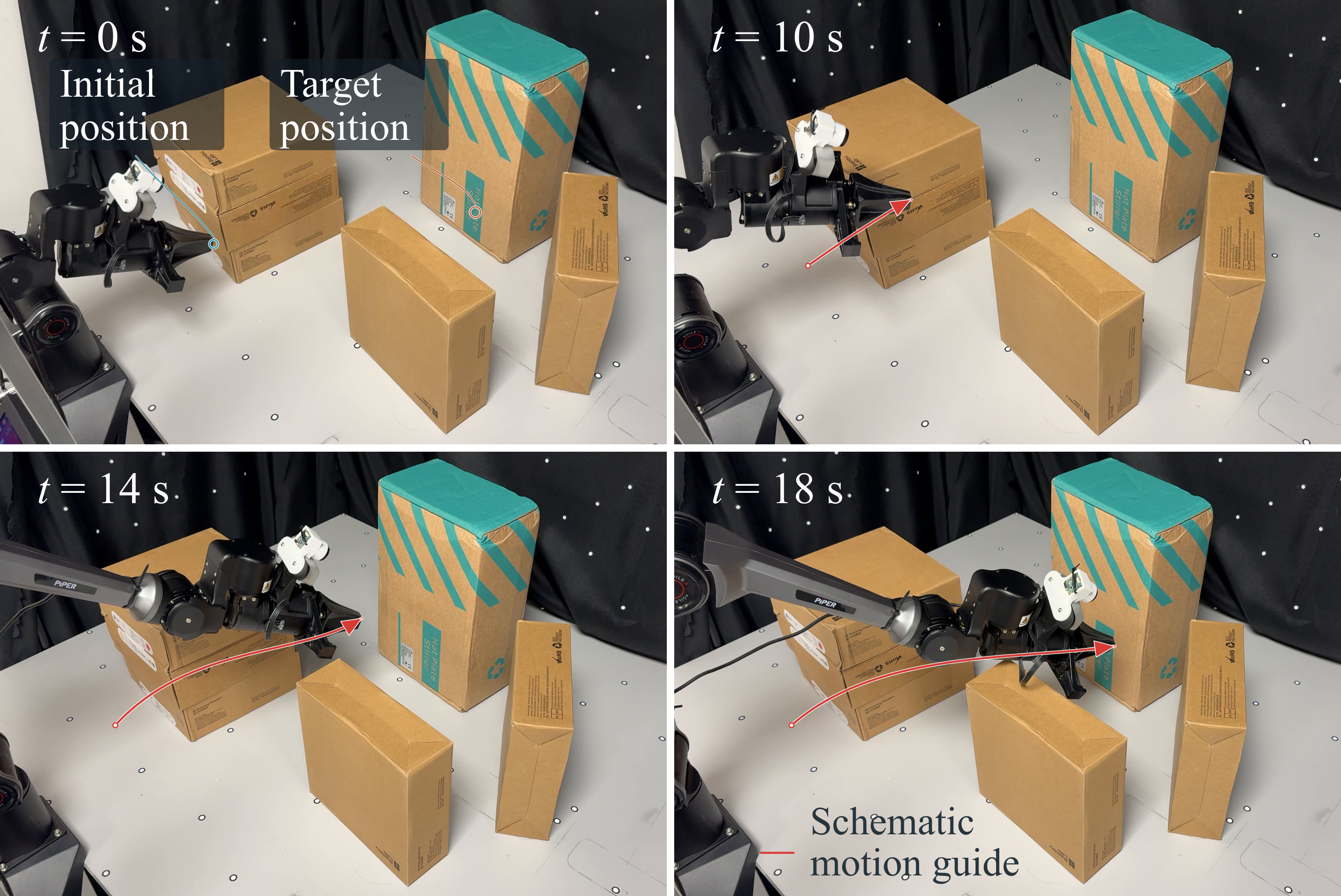}
  \caption{Physical PIPER motion among cardboard obstacles at $t=0,10,14,18$~s.}
  \label{fig:hardware-demo}
\end{figure}

A physical PIPER arm and gripper performed a tabletop reaching demonstration among
stationary cardboard obstacles (Fig.~\ref{fig:hardware-demo}). A two-box stack and three
individual boxes formed four obstacle groups of different heights. The arm was to extend
from its initial configuration toward the marked target region.

Four frames from a single recording show the motion at $t=0,10,14,18$~s. The arm was
initially retracted, with the gripper to the left of the boxes. The intermediate frames
captured the gripper lifting at $t=10$~s and advancing farther into the arrangement at
$t=14$~s. By $t=18$~s, the arm was more extended and the gripper was closer to the marked
target.

The sequence provides a qualitative example of reaching among physical obstacles.
The certification decisions, clearance margin and computational comparisons reported above
refer to the simulation study.

\Needspace{4\baselineskip}
\section{Discussion}
Most of the savings arise from portions of the motion that are far from any obstacle.
\method{} certifies these portions with a few long intervals and concentrates its evaluations
where clearance is small. The shelf trajectory of Fig.~\ref{fig:overview}(d) illustrates this
pattern: the early part of the motion passes at depth three or four, and deeper refinement is
confined to the later part, as the arm approaches the shelf. No interval reaches depth nine,
and \method{} needs 105 evaluations where fixed-fine needs 512. Such portions are shorter in
the medium-window and easy-shelf scenes, which show the highest evaluation ratios and the
smallest speedups, $3.14\times$ and $4.18\times$.

\method{} and fixed-fine left the same 27 sampled-clear trajectories uncertified, so these
rejections come from the shared certificate model, not from adaptive refinement. The model is
conservative in three respects: the capsules over-approximate the link meshes, the motion
radius sums worst-case contributions from every ancestor joint, and the 10~mm margin is added
to both. Deeper refinement shrinks only the motion term. As $|I|\rightarrow0$, the motion
radius $b_\ell(I)$ vanishes, but the gap between each mesh and its enclosing capsule does not.

Because \method{} takes the trajectory as given, it can be applied downstream of any planner
that outputs joint positions and velocities. Algorithm~\ref{alg:larc} is also independent of
how the envelope is constructed, so a tighter enclosure can be used within the same tree
without further changes.

\section{Conclusion}
\method{} determines whether every link of a planned trajectory stays clear of the obstacles
for the entire motion, and it splits a time interval only when its clearance test is
inconclusive. With midpoint capsules and exact Hermite speed bounds, the certificate holds in
continuous time: no link of a certified trajectory comes within the margin of a static
obstacle. On the simulated PIPER dataset, \method{} reached the same decisions as fixed-fine refinement
with about a quarter of the interval evaluations and roughly a tenth of the runtime, and it
certified none of the trajectories in which the audit found a collision. About a fifth of the
sampled-clear trajectories were left uncertified; we regard tighter link enclosures as the
most direct means of certifying more of them. A physical PIPER demonstration illustrates
reaching among cardboard obstacles. Future work will also extend the bounds to
tracking error, moving obstacles, and self-collision.

\bibliographystyle{IEEEtran}
\bibliography{Reference}

@misc{holmesReachableSetsSafe2020a,
  author        = {Patrick Holmes and Shreyas Kousik and Bohao Zhang and Daphna Raz and Corina Barbalata and Matthew Johnson-Roberson and Ram Vasudevan},
  title         = {Reachable Sets for Safe, Real-Time Manipulator Trajectory Design},
  year          = {2020},
  eprint        = {2002.01591},
  archivePrefix = {arXiv},
  primaryClass  = {cs.RO},
  note = {arXiv preprint arXiv:2002.01591}
}

@misc{michauxSafePlanningArticulated2024a,
  author        = {Jonathan Michaux and Adam Li and Qingyi Chen and Che Chen and Bohao Zhang and Ram Vasudevan},
  title         = {Safe Planning for Articulated Robots Using Reachability-based Obstacle Avoidance With Spheres},
  year          = {2024},
  eprint        = {2402.08857},
  archivePrefix = {arXiv},
  primaryClass  = {cs.RO},
  note = {arXiv preprint arXiv:2402.08857}
}

@inproceedings{kwonConformalizedReachableSets2025,
  author    = {Yongseok Kwon and Jonathan Michaux and Seth Isaacson and Bohao Zhang and Matthew Ejakov and Katherine A. Skinner and Ram Vasudevan},
  title     = {Conformalized Reachable Sets for Obstacle Avoidance with Spheres},
  booktitle = {2025 IEEE International Conference on Robotics and Automation (ICRA)},
  year      = {2025},
  pages     = {12877--12884},
}

@inproceedings{rudorferRM4DCombinedReachability2025,
  author    = {Martin Rudorfer},
  title     = {{RM4D}: A Combined Reachability and Inverse Reachability Map for Common 6-/7-Axis Robot Arms by Dimensionality Reduction to 4D},
  booktitle = {2025 IEEE International Conference on Robotics and Automation (ICRA)},
  year      = {2025},
  pages     = {7689--7695},
}

@inproceedings{makhalReuleauxRobotBase2018,
  author    = {Abhijit Makhal and Alex K. Goins},
  title     = {Reuleaux: Robot Base Placement by Reachability Analysis},
  booktitle = {2018 Second IEEE International Conference on Robotic Computing (IRC)},
  year      = {2018},
  pages     = {137--142},
}

@inproceedings{kimLearningReachableManifold2021a,
  author    = {Seungsu Kim and Julien Perez},
  title     = {Learning Reachable Manifold and Inverse Mapping for a Redundant Robot Manipulator},
  booktitle = {2021 IEEE International Conference on Robotics and Automation (ICRA)},
  year      = {2021},
  pages     = {4731--4737},
}

@article{cavelliModelingReachabilitySpace2025b,
  author  = {Rosario Francesco Cavelli and Pangcheng David Cen Cheng and Marina Indri},
  title   = {Modeling the Reachability Space of Robotic Manipulators through Ellipsoid Equations},
  journal = {Journal of Intelligent \& Robotic Systems},
  year    = {2025},
  volume  = {111},
  pages   = {90},
}

@inproceedings{stulpCombiningAnalysisImitation2009,
  author    = {Freek Stulp and Andreas Fedrizzi and Franziska Zacharias and Moritz Tenorth and Jan Bandouch and Michael Beetz},
  title     = {Combining Analysis, Imitation, and Experience-Based Learning to Acquire a Concept of Reachability in Robot Mobile Manipulation},
  booktitle = {Proc. IEEE-RAS Int. Conf. Humanoid Robots},
  pages     = {161--167},
  year      = {2009},
}

@inproceedings{zachariasUsingModelReachable2009,
  author    = {Franziska Zacharias and Wolfgang Sepp and Christoph Borst and Gerd Hirzinger},
  title     = {Using a Model of the Reachable Workspace to Position Mobile Manipulators for 3-D Trajectories},
  booktitle = {Proc. IEEE-RAS Int. Conf. Humanoid Robots},
  pages     = {55--61},
  year      = {2009},
}

@inproceedings{louLearningGenerate6DoF2020,
  author    = {Xibai Lou and Yang Yang and Changhyun Choi},
  title     = {Learning to Generate 6-{DoF} Grasp Poses with Reachability Awareness},
  booktitle = {Proc. IEEE Int. Conf. Robot. Autom.},
  pages     = {1532--1538},
  year      = {2020},
}

@inproceedings{akinolaDynamicGraspingReachability2021,
  author    = {Iretiayo Akinola and Jingxi Xu and Shuran Song and Peter K. Allen},
  title     = {Dynamic Grasping with Reachability and Motion Awareness},
  booktitle = {Proc. IEEE/RSJ Int. Conf. Intell. Robots Syst.},
  pages     = {9422--9429},
  year      = {2021},
}

@article{zhuStructuralAnalysisDesign2025a,
  author  = {Xin Zhu and Haozhou Liu and Qingqing Li and Zhaoyang Cai and Xuechao Chen and Zhangguo Yu and Qiang Huang and Abderrahmane Kheddar},
  title   = {Structural Analysis and Design of Humanoid Arms From Human Arm Reachable Workspace},
  journal = {IEEE Robot. Autom. Lett.},
  volume  = {10},
  number  = {2},
  pages   = {1233--1240},
  year    = {2025},
}

@article{fengPredictiveReachabilityEmbodiment2025,
  author  = {Xiaoxu Feng and Takato Horii and Takayuki Nagai},
  title   = {Predictive Reachability for Embodiment Selection in Mobile Manipulation Behaviors},
  journal = {IEEE Robot. Autom. Lett.},
  volume  = {10},
  number  = {3},
  pages   = {2966--2973},
  year    = {2025},
}

@article{mengaldoConciseGuideModelling2022,
  author  = {Gianmarco Mengaldo and Federico Renda and Steven L. Brunton and Moritz B{\"a}cher and Marcello Calisti and Christian Duriez and Gregory S. Chirikjian and Cecilia Laschi},
  title   = {A Concise Guide to Modelling the Physics of Embodied Intelligence in Soft Robotics},
  journal = {Nature Reviews Physics},
  volume  = {4},
  pages   = {595--610},
  year    = {2022},
}

@article{kappassovTactileSensingDexterous2015,
  author  = {Zhanat Kappassov and Juan-Antonio Corrales and V{\'e}ronique Perdereau},
  title   = {Tactile Sensing in Dexterous Robot Hands---Review},
  journal = {Robotics and Autonomous Systems},
  volume  = {74},
  pages   = {195--220},
  year    = {2015},
}

@article{andrussowMinsightFingertipSized2023,
  author  = {Iris Andrussow and Huanbo Sun and Katherine J. Kuchenbecker and Georg Martius},
  title   = {{Minsight}: A Fingertip-Sized Vision-Based Tactile Sensor for Robotic Manipulation},
  journal = {Advanced Intelligent Systems},
  volume  = {5},
  number  = {8},
  pages   = {2300042},
  year    = {2023},
}

@article{deimelNovelTypeCompliant2016a,
  author  = {Raphael Deimel and Oliver Brock},
  title   = {A Novel Type of Compliant and Underactuated Robotic Hand for Dexterous Grasping},
  journal = {The International Journal of Robotics Research},
  volume  = {35},
  number  = {1--3},
  pages   = {161--185},
  year    = {2016},
}

@misc{wuDexLinkHandCompact2026,
  author        = {Hao Wu and Yanzhe Wang and Yu Feng and Jian Liu and Jihao Li and Jianshu Zhou and Huixu Dong},
  title         = {{DexLink} Hand: A Compact, Affordable, 16-{DOF} Linkage-Driven Hand with Human-Like Dexterity},
  year          = {2026},
  note          = {arXiv preprint arXiv:2606.17418},
  eprint        = {2606.17418},
  archiveprefix = {arXiv},
  primaryclass  = {cs.RO},
}

@misc{wuSyLinkHandSynergyInspired2026,
  author        = {Hao Wu and Yanzhe Wang and Yu Feng and Yitong Li and Jingxiang Guo and Jian Liu and Jianshu Zhou},
  title         = {{SyLink} Hand: A Synergy-Inspired Linkage-Driven Anthropomorphic Hand for Human-Like Dexterity},
  year          = {2026},
  note          = {arXiv preprint arXiv:2606.14250},
  eprint        = {2606.14250},
  archiveprefix = {arXiv},
  primaryclass  = {cs.RO},
}

@misc{wuSpectraTacCompactCameraFree2026,
  author        = {H. Wu and H. Guo and Y. Feng and Y. Wang and Y. Wang and J. Zhou},
  title         = {{SpectraTac}: A Compact Camera-Free Optical Tactile Sensor with Distributed Color Sensing},
  year          = {2026},
  note          = {arXiv preprint arXiv:2608.30368},
  eprint        = {2608.30368},
  archiveprefix = {arXiv},
}

@article{fengSoftMechanoluminescentSkin2026,
  author  = {Yu Feng and Qiaojiao Wang and Yehui Liu and Jiankun Li and Senlin Hou and Xiaomeng Yang and Ziyi Li and Yanji Yi and Meng Chen and Guanglie Zhang and Hao Sun and Wen Jung Li},
  title   = {A Soft Mechanoluminescent Skin for High-Resolution Optical Tactile Sensing in Human--Machine Interaction},
  journal = {Advanced Science},
  volume  = {13},
  number  = {42},
  pages   = {e75507},
  year    = {2026},
}

@article{fengBioinspiredSoftMagnetociliary2026,
  author  = {Yu Feng and Yuzhao Zhang and Yehui Liu and Guanglie Zhang and Xinge Yu and Jianshu Zhou and Wen Jung Li},
  title   = {A Bioinspired Soft Magnetociliary Tactile Sensor for Direction-Sensitive Interaction Sensing},
  journal = {IEEE Trans. Instrum. Meas.},
  volume  = {75},
  pages   = {9531313},
  year    = {2026},
}

@article{zhouEverythingGraspingGripperUniversal2025,
  author  = {Jianshu Zhou and Jing Shu and Tianle Pan and Puchen Zhu and Jiajun An and Huayu Zhang and Junda Huang and Upinder Kaur and Xin Ma and Masayoshi Tomizuka},
  title   = {Everything-Grasping Gripper: A Universal Gripper with Synergistic Suction--Grasping Capabilities for Cross-Scale and Cross-State Manipulation},
  journal = {Soft Robotics},
  volume  = {13},
  number  = {2},
  pages   = {178--189},
  year    = {2026},
}

@article{zhouDexterousCompliantDexCo2025b,
  author  = {Jianshu Zhou and Junda Huang and Qi Dou and Pieter Abbeel and Yunhui Liu},
  title   = {A Dexterous and Compliant ({DexCo}) Hand Based on Soft Hydraulic Actuation for Human-Inspired Fine In-Hand Manipulation},
  journal = {IEEE Trans. Robot.},
  volume  = {41},
  pages   = {666--686},
  year    = {2025},
}

@misc{fengMagCilia2026,
  author = {Feng, Yu and Wu, Hao and Guo, Haotian and Liu, Haoming and Su, William and Guo, Jingxiang and Li, Jiankun and Tomizuka, Masayoshi and Li, Wen Jung and Zhou, Jianshu},
  title = {{MagCilia}: A Compact Magnetociliary Tactile Sensor with {3D} Force Sensing for Robotic Contact Perception and Grasping Feedback},
  year = {2026},
  eprint = {2610.09536},
  archivePrefix = {arXiv},
  primaryClass = {cs.RO},
  note = {arXiv preprint arXiv:2610.09536}
}

@article{zhouPrismaticBending2025,
  author = {Zhou, Jianshu and Huang, Junda and Liang, Boyuan and Zhang, Xiang and Ma, Xin and Tomizuka, Masayoshi},
  title = {Prismatic-Bending Transformable ({PBT}) Joint for a Modular, Foldable Manipulator With Enhanced Reachability and Dexterity},
  journal = {IEEE/ASME Transactions on Mechatronics},
  year = {2025},
  volume = {30},
  number = {4},
  pages = {2955--2963},
}

@misc{zhouGlobalLocal2025,
  author = {Zhou, Jianshu and Liang, Boyuan and Huang, Junda and Zhang, Ian and Tomizuka, Masayoshi},
  title = {Global-Local Interface for On-Demand Teleoperation},
  year = {2025},
  eprint = {2502.09960},
  archivePrefix = {arXiv},
  primaryClass = {cs.RO},
  note = {arXiv preprint arXiv:2502.09960}
}

@article{huangDIHTele2025,
  author = {Huang, Junda and Chen, Kai and Zhou, Jianshu and Lin, Xingyu and Abbeel, Pieter and Dou, Qi and Liu, Yunhui},
  title = {{DIH-Tele}: Dexterous In-Hand Teleoperation Framework for Learning Multiobjects Manipulation With Tactile Sensing},
  journal = {IEEE/ASME Transactions on Mechatronics},
  year = {2025},
  volume = {30},
  number = {5},
  pages = {3840--3851}
}

@article{caoOrigamiGripper2024,
  author = {Cao, Hanwen and Zhou, Jianshu and Chen, Kai and He, Qiguang and Dou, Qi and Liu, Yun-Hui},
  title = {Design and Optimization of an Origami Gripper for Versatile Grasping and Manipulation},
  journal = {Advanced Intelligent Systems},
  year = {2024},
  volume = {6},
  number = {12},
  pages = {2400271}
}

@article{zhouBioinspiredSoftWrist2023,
  author = {Zhou, Jianshu and Cao, Hanwen and Chen, Wei and Cheng, Shing Shin and Liu, Yun-Hui},
  title = {Bioinspired Soft Wrist Based on Multicable Jamming With Hybrid Motion and Stiffness Control for Dexterous Manipulation},
  journal = {IEEE/ASME Transactions on Mechatronics},
  year = {2023},
  volume = {28},
  number = {3},
  pages = {1256--1267}
}

\end{document}